\documentclass[10pt,twocolumn,letterpaper]{article}

\usepackage{iccv}
\usepackage{times}
\usepackage{epsfig}
\usepackage{graphicx}
\usepackage{amsmath}
\usepackage{amssymb}
\usepackage{pifont}

\usepackage[breaklinks=true,bookmarks=false]{hyperref}

\iccvfinalcopy 

\def\iccvPaperID{****} 

\ificcvfinal\fi

\begin{document}

\title{Occ$^2$Net: Robust Image Matching Based on 3D Occupancy Estimation for Occluded Regions}

\author{Miao Fan \qquad Mingrui Chen \qquad Chen Hu \qquad Shuchang Zhou\\
MEGVII Technology\\
{\tt\small \{fanmiao02, chenmingrui, huchen, zsc\}@megvii.com}
}

\maketitle

\ificcvfinal\thispagestyle{empty}\fi

\begin{abstract}
Image matching is a fundamental and critical task in various visual applications, such as Simultaneous Localization and Mapping (SLAM) and image retrieval, which require accurate pose estimation. However, most existing methods ignore the occlusion relations between objects caused by camera motion and scene structure.
In this paper, we propose Occ$^2$Net, a novel image matching method that models occlusion relations using 3D occupancy and infers matching points in occluded regions. 
Thanks to the inductive bias encoded in the Occupancy Estimation (OE) module, it greatly simplifies bootstrapping of a multi-view consistent 3D representation that can then integrate information from multiple views. Together with an Occlusion-Aware (OA) module, it incorporates attention layers and rotation alignment to enable matching between occluded and visible points. 
We evaluate our method on both real-world and simulated datasets and demonstrate its superior performance over state-of-the-art methods on several metrics, especially in occlusion scenarios.
\end{abstract}

\section{Introduction}
\label{sec:intro}
Image matching is a fundamental and critical task in various visual applications, such as SLAM and image retrieval. It aims to identify and correspond to the same or similar structure/contents from two or more images. Image matching can be divided into two categories: feature-based methods \cite{rublee2011orb, lowe2004distinctive,detone2018superpoint, sarlin2020superglue} and dense methods \cite{sun2021loftr, jiang2021cotr}. Feature-based methods extract sparse keypoints and descriptors from images and match them based on similarity metrics, while dense methods estimate dense correspondences between pixels or patches of images.

\begin{figure}[t]
 \centering
  \includegraphics[width=1.0\linewidth]{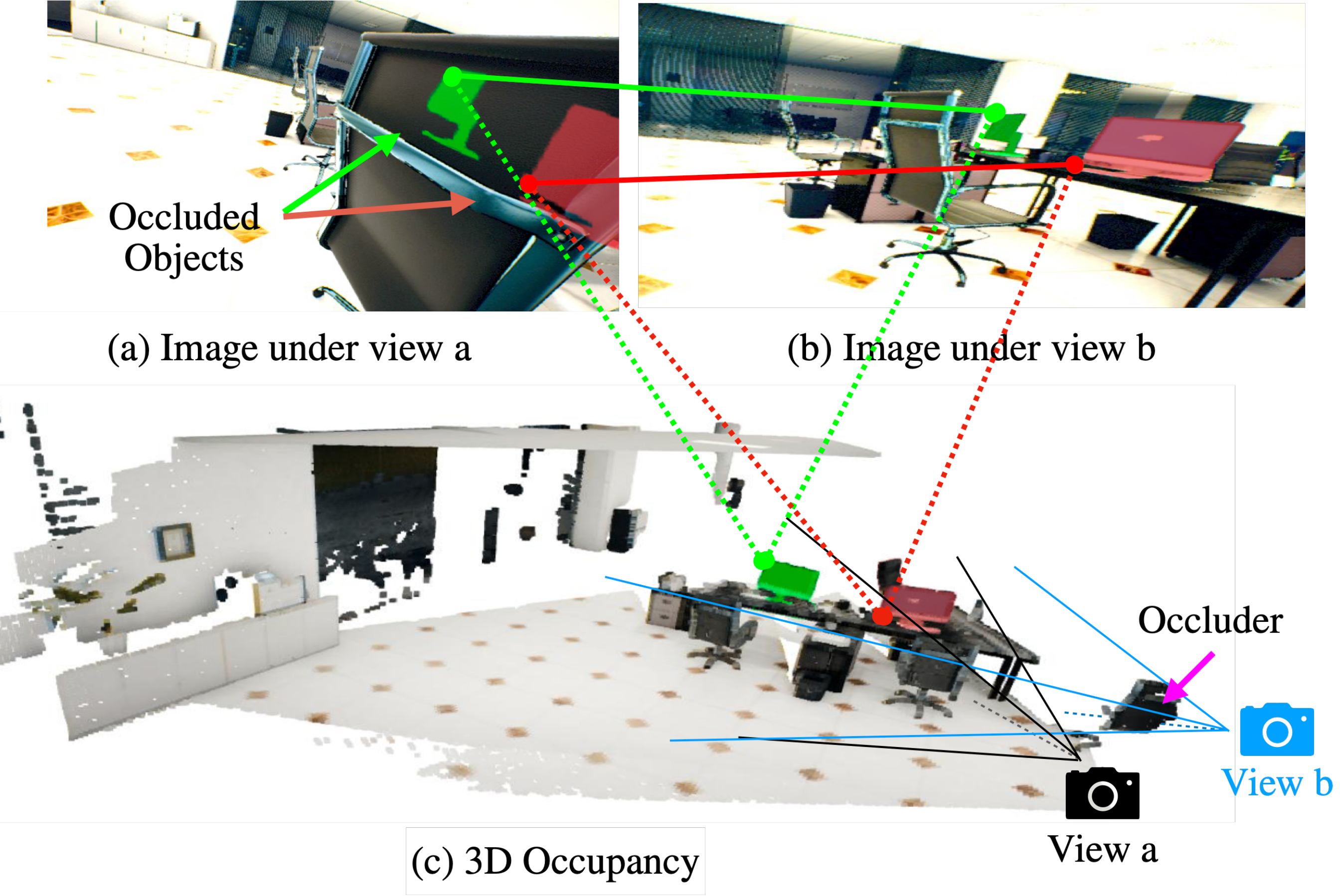}
  \caption{\textbf{Schematic diagram of the  Occ$^2$Net.} (a) and (b) are images taken from different viewpoints, while (c) shows the matching process for occluded regions. In (c), two monitors are shown with green and red masks indicating areas that are visible in (b) but occluded in (a). By using Occ$^2$Net to extract consistent occupancy features and match them between (a) and (b), the monitors that are occluded in (a) can still be matched in (b), thus enabling Occ$^2$Net to have the ability to perform matching under occlusion.}
  \label{fig: question}
\end{figure}

However, both types of methods struggle with occlusion scenarios, which are frequent in real-world environments. Fig. \ref{fig: question} shows an example of these challenges. The two images have a large disparity due to camera motion. Although there are considerable overlapping regions, the large disparity causes occlusion, which significantly reduces the number of visible matching pairs. Moreover, in this example, both the ground and wall in the scene have low texture, and large areas are visible in image (b) but occluded in image (a), such as the two distinguishable monitors marked as green and red. These factors make it difficult for existing algorithms to extract enough matching pairs for camera pose estimation. Scenarios like these are common in indoor navigation or autonomous driving. To tackle these problems, we propose a novel image matching method called Occ$^2$Net, which matches not only the visible point pairs but also the occluded points and the visible points.

Based on this observation, we design Occ$^2$Net to match 3D points. Following NeRF \cite{mildenhall2021nerf}, we treat each pixel as a ray emitted from the corresponding camera. NeRF \cite{mildenhall2021nerf} obtains 3D points along the ray by sampling at equal intervals and learns their information by differentiable rendering. However, in the matching algorithm, we do not have pose information at inference time, so we simplify the sampling along the ray to two points: one visible point and one occluded point. At training time, we use ground truth depths and poses to reproject and determine whether a 3D point is occluded or visible.

Based on these simplifications, Occ$^2$Net extends the matching between visible-visible points to matching between visible-occluded points. To achieve this goal, we use a 3D Occupancy Estimation (OE) Module, which greatly simplifies the multi-view 3D representation, following \cite{cao2022monoscene, miao2023occdepth}. Due to the difficulty of 3D matching, the large GPU memory for occupancy and the error in the occupancy estimation, we do not use the occupancy of the whole image to estimate the matching. We use a coarse-to-fine structure instead. The Occlusion-Aware (OA) module is used in the coarse stage to obtain the matching between patches and the OE module is used to obtain the fine matching points.

 We evaluate our proposed method on two datasets: ScanNet \cite{dai2017scannet} and TartanAir \cite{wang2020tartanair}, containing real-world and simulated scenes with various degrees of occlusion. We compare our method with several state-of-the-art feature-based and dense methods on several metrics. Experiments show that our method achieves superior accuracy on both datasets, outperforming existing methods by large margins. Moreover, our method demonstrates robustness and efficiency under occlusion scenarios.

 In conclusion, we propose an image matching algorithm that is aware of occluded points and outperforms state-of-the-art methods on both real-world and synthetic datasets. Specifically, our contributions are: 
 \begin{itemize} 
 \item  We propose Occ$^2$Net, a novel occlusion-aware image matching algorithm that uses 3D occupancy to model the occlusion relations between objects and infer the location of matching points in occluded regions.
 \item We combine an Occupancy Estimation (OE) module with an Occlusion-Aware (OA) module to enable visible-occluded matching using a coarse-to-fine structure with occupancy estimation.
 \item Our experiments show  Occ$^2$Net achieves state-of-the-art pose estimation accuracy on both the real-world dataset ScanNet and the simulated dataset TartanAir. 
 \end{itemize}


 \begin{figure*}[t] 
  \centering \includegraphics[width=1.0\linewidth]{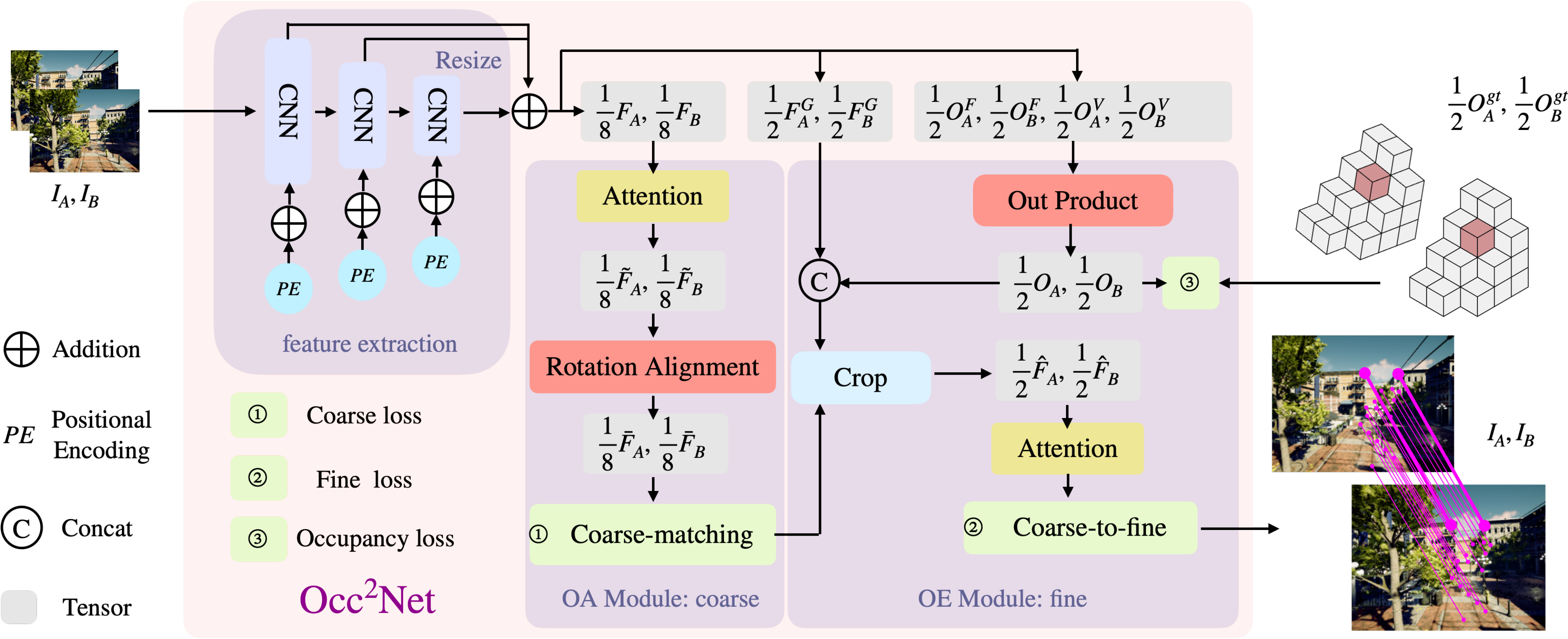}
  \caption{\textbf{Occ$^2$Net framework.} We first extract features, then apply OA Module to obtain coarse matching patches, and finally combine fine features with 3D occupancy estimates in OE Module for fine matching.}
  \label{fig:ppl}
\end{figure*}

\section{Related Work}
\label{sec:related}

\textbf{Traditional keypoint detection and feature description methods} are widely used for image matching. There are three basic steps in traditional method: interest points detection, feature description, and feature matching. In order to perform better image matching, before the era of deep learning, computer vision researchers have designed many hand-crafted feature descriptors, e.g. SIFT \cite{lowe2004distinctive}, SURF \cite{bay2008speeded}, ORB \cite{rublee2011orb}, DISK \cite{tyszkiewicz2020disk}, etc., which are invariant under changes of lighting, exposure or camera movement. Feature detection and description are widely used in modern SLAM systems. For example, the ORB-SLAM \cite{mur2015orb,mur2017orb,campos2021orb} series is designed based on ORB \cite{rublee2011orb} descriptors. Feature matching is the process of calculating the distances between feature vectors. Commonly used metrics defined on feature vectors include Euclidean distance \cite{danielsson1980euclidean}, Hamming distance \cite{norouzi2012hamming}, and cosine distance \cite{senoussaoui2013study}. However, they have limited ability to capture semantic information and handle large viewpoint changes.

\textbf{Learnable detection and matching methods based on neural networks} have emerged with the development of deep learning algorithms. LIFT \cite{yi2016lift} and MagicPoint \cite{detone2017toward} are the first learning-based methods. Then the SuperPoint \cite{detone2018superpoint}, a self-supervised training method, came to everyone's eyes. SuperPoint \cite{detone2018superpoint} is built upon MagicPoint \cite{detone2017toward} and uses Homographic Adaptation to generate pseudo-ground truth matching for supervision. The above-mentioned methods use the nearest neighbor searching to find matches among the detected interest points. SuperGlue \cite{sarlin2020superglue} leverages a graph neural network \cite{shi2020point}(GNN) to learn priors of transformations and regularities of the 3D world and achieves state-of-the-art results on pose estimation task. The SuperPoint \cite{detone2018superpoint} + SuperGlue \cite{sarlin2020superglue} is widely used in SLAM systems \cite{qin2020sp,xu2020cnn} and inspired a series of detector-based deep learning approaches \cite{li2021deep,zhong2018detect,yi2018learning,zhang2019learning,luo2019contextdesc,lowe2004distinctive}. D2-Net \cite{dusmanu2019d2} jointly trains the feature detectors and descriptors, and guides the detection of interest points through the descriptors. These methods learn stronger semantic features from a large amount of data and achieve impressive results. However, they still rely on keypoint detection as an intermediate step, which may introduce error or miss matches.

\textbf{End-to-end detector-free image matching} algorithms have been proposed in the past two years. Detector-free methods \cite{rocco2020efficient,li2020dual} skip the feature detector phase and directly match dense pixels. NCNet \cite{rocco2020ncnet} proposes an end-to-end CNN that outputs matches by analyzing neighborhood consensus patterns from of all possible correspondences. Recently, Transformers \cite{vaswani2017attention} are getting more attention in computer vision tasks. COTR \cite{jiang2021cotr} is a Transformer-based method using zooming in method to obtain more accurate matching. LoFTR \cite{sun2021loftr} obtains confidence matrices for coarse matching, and refines matching points positions at fine level, both steps using self and cross attention layers in Transformer. QuadTree LoFTR \cite{tang2022quadtree} is a follow-up method on the LoFTR \cite{sun2021loftr} framework, which uses a tree data structure, in which each internal node has exactly four children, to improve matching accuracy. ASpanFormer \cite{chen2022aspanformer} develops an attention operation which adjusts attention span according to the computed flow maps and the adaptive sampling grid size. These methods can produce more accurate matching pairs by directly learning dense correspondences between images without keypoint detection. However, these methods are still not robust enough for some practical applications such as SLAM, especially when part of objects are occluded in the corresponding images.

\textbf{Occlusion-aware algorithms} The inpainting task and the amodal perception task have some relevance to our occlusion-aware image matching. \cite{mohan2022amodal} propose a novel task named amodal panoptic segmentation. The goal of this task is to simultaneously predict the pixel-wise semantic segmentation labels of the visible regions of stuff classes and the instance segmentation labels of both the visible and occluded regions of thing classes.  \cite{back2022unseen} propose a Hierarchical Occlusion Modeling (HOM) scheme to reason about the occlusion by assigning a hierarchy to feature fusion and prediction order. \cite{lee2022instance} mainly consider the relationship between occlusion order and depth order.  \cite{ao2022image} provides an intuitive understanding of the research hotspots, key technologies, and future trends in the field of image amodal completion. Amodal completion is still difficult for complex indoor and outdoor scenes. 

\cite{li2022mat,quan2022image,dong2022incremental,zeng2022aggregated,zhang2022gan} have achieved impressive results in image inpainting task.  These algorithms, however, require additional mask inputs and complement the missing textures based on surrounding information. In the matching task, the occluded locations have foreground objects, so we need to rely only on the visible partly-occluded objects and not on the texture of the foreground objects to complete the occlusion. Since lack of mask to identify occluded regions, we apply positional encoding at multiple scales across the entire image.

Visual Correspondence Hallucination \cite{germain2021visual} is able to output the peaked probability distribution over the matching location, regardless of the correspondent being visible, occluded, or outside the field of view. This method is mainly for visualization purpose but cannot improve the accuracy for pose estimation by our experiments.

\textbf{In this paper}, the occlusion-aware algorithm is combined with the image matching algorithm and the 3D occupancy is estimated to infer the location of the occluded points to achieve robust image matching.

\section{Methods}
\label{sec:method}
An overview of our Occ$^2$Net is presented in Fig.\ref{fig:ppl}, which aids in matching under occlusion by implicitly modeling the object-occlusion relationship. Our method has been extensively tested on various scenarios of visible-visible and visible-occluded and achieves comparable results with state-of-the-art methods. However, in this section, our focus is mainly on the challenging visible-occluded matching.

Our method consists of three modules. Firstly, we extract multi-scale features (Sec. \ref{sec:feature_extraction}). Then we use OA Module (Sec. \ref{sec:OA Module}) to perform coarse matching based on the feature extraction output. Finally, we propose the OE Module (Sec. \ref{sec:OE Module}) to infer the location of exact matches of each coarse patch. 

\subsection{Pipeline Overview}
\label{sec:ppl}
Our method takes two input images $I_A$ and $I_B$ of size $(H,W)$ as shown in Fig. \ref{fig:ppl}. It extracts multi-layer features and adds multi-scale positional encodings $PE$, and then outputs two types of features: coarse-level features $F$ of $\frac{1}{8}$ resolution and fine-level features $F^G$ of $\frac{1}{2}$ resolution. The feature extraction also outputs feature tensors $O^F$ and $O^V$ for 3D occupancy estimation. The OA Module first uses attention components to combine features $F$ and obtain $\tilde{F}$, which have the same dimension as $F$. Then Rotation Alignment is applied on each feature to rotate the corresponding patch feature, as shown in Fig. \ref{fig: rotation_method}. After Rotation Alignment, we get features $\bar{F}$. The OE Module first estimates the 3D occupancy $O$ of each image by $O^F$ and $O^V$ of $\frac{1}{2}$ resolution. After that, OE Module obtains $\hat{F}$, the combined feature of the 3D occupancy $O$  and fine features  $F^G$  of the corresponding locations, based on the coarse matching results. Finally, the location of the matching points are deduced using attention. 

\subsection{Feature Extraction}
\label{sec:feature_extraction}
Our feature extraction has three main functions: coarse feature $F$ extraction, fine feature $F^G$ extraction, and generating tensors $O^F$ and $O^V$ for 3D occupancy estimation. For matching both visible and occluded points, we use a pyramid structure \cite{ma2018shufflenet} with a large receptive field and add positional encoding \cite{wu2021rethinking} for different scales.

The pyramid structure helps recognizing the same object of different scales due to different camera distances. In order to match occluded patches with visible patches successfully, we enlarge the receptive field of the pyramid structure and use more surrounding information for matching. As Fig. \ref{fig: question} shows, the patches containing the chair in the images are different in the chair part but are very similar in the surrounding region. It is a common phenomenon that the image features surrounding matched occluded regions and visible regions are similar. We also add positional encoding of different scales, which enriches the features with position information for correctly inferring the offsets of the occluded points from the above-mentioned similar surrounding region.

\subsection{OA Module}
\label{sec:OA Module}

We design the OA Module to achieve coarse matching. The OA Module mainly consists of two parts: attention component and Rotation Alignment. The attention component deepens the understanding of the structure information of the whole image through self-attention and cross-attention, making the features more conducive to coarse matching. The Rotation Alignment aims to better adapt to different rotations between different viewpoints, which makes both visible and occluded patches easier to match.

\textbf{Rotation Alignment} The Rotation Alignment selects features with suitable rotation angles. After the attention component, each feature retains the patch information of its receptive field. The features of the left image after a suitable rotation are closer to those of the right image, which further is aware of occlusion and make patch matching easier.

\begin{figure}[t]
 \centering
  \includegraphics[width=1.0\linewidth]{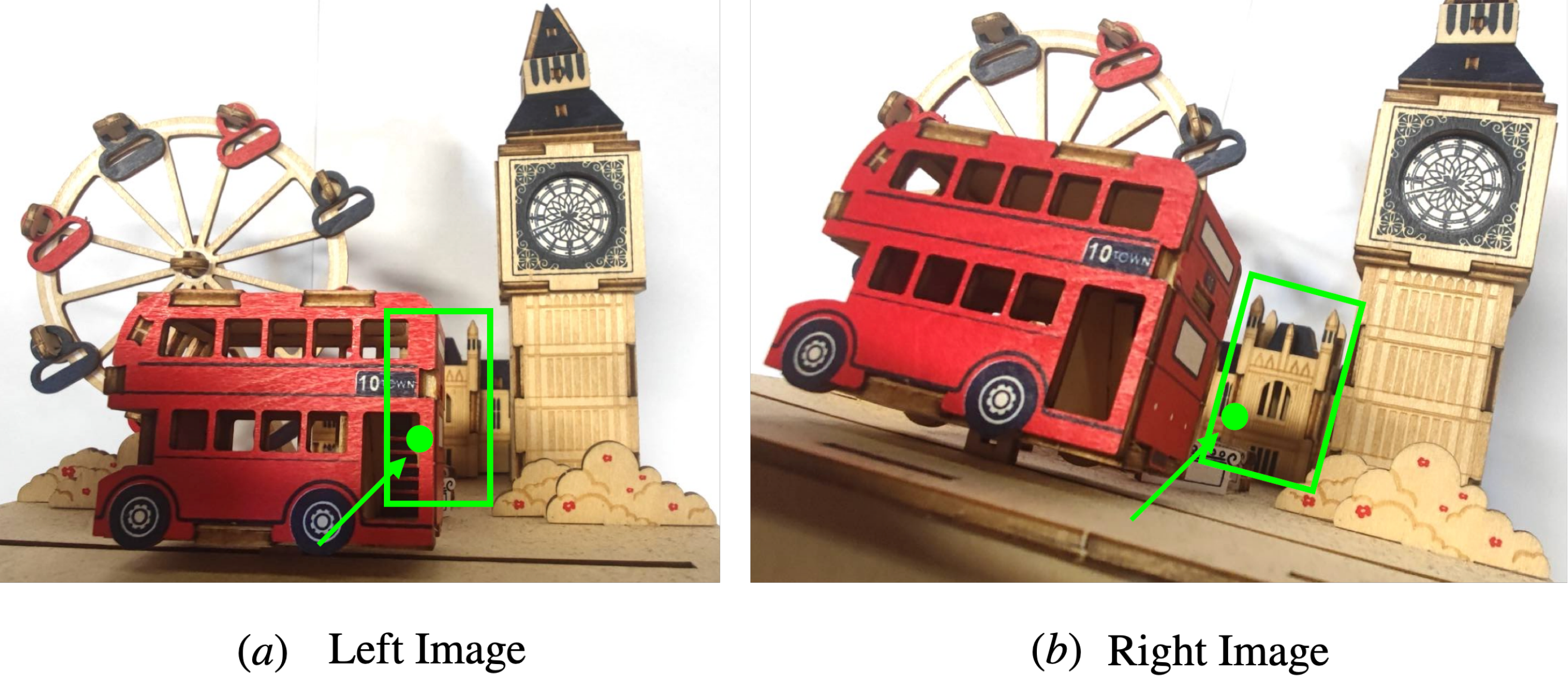}
  \caption{\textbf{An example to illustrate the role of the Rotation Alignment of OA Module.} We want to match the points indicated by the arrow in Figure (a) and (b). The points in Figure (a) are occluded and visible in Figure (b). In the coarse matching phase, we need to match the two green patches in Figure. However, due to the occlusion region, the features of the two patches will differ slightly. The goal of the Rotation Alignment is to increase the similarity of the features of the two patches.}
  \label{fig:rotation}
\end{figure}

For example, as Fig. \ref{fig:rotation} shows, we want to match the points indicated by the arrows in Figure (a) and (b). The points are occluded in Figure (a) but visible in Figure (b). In the coarse matching phase, we need to match the two green patches. However, due to the occlusion region, the features of the two patches will differ slightly. Even though we have expanded the receptive field in feature extraction and introduced more surrounding information, we cannot remove the effect of occlusion. The goal of the Rotation Alignment is to increase the feature similarity of the two patches.

\begin{figure}[t]  
  \centering
  \includegraphics[width=1.0\linewidth]{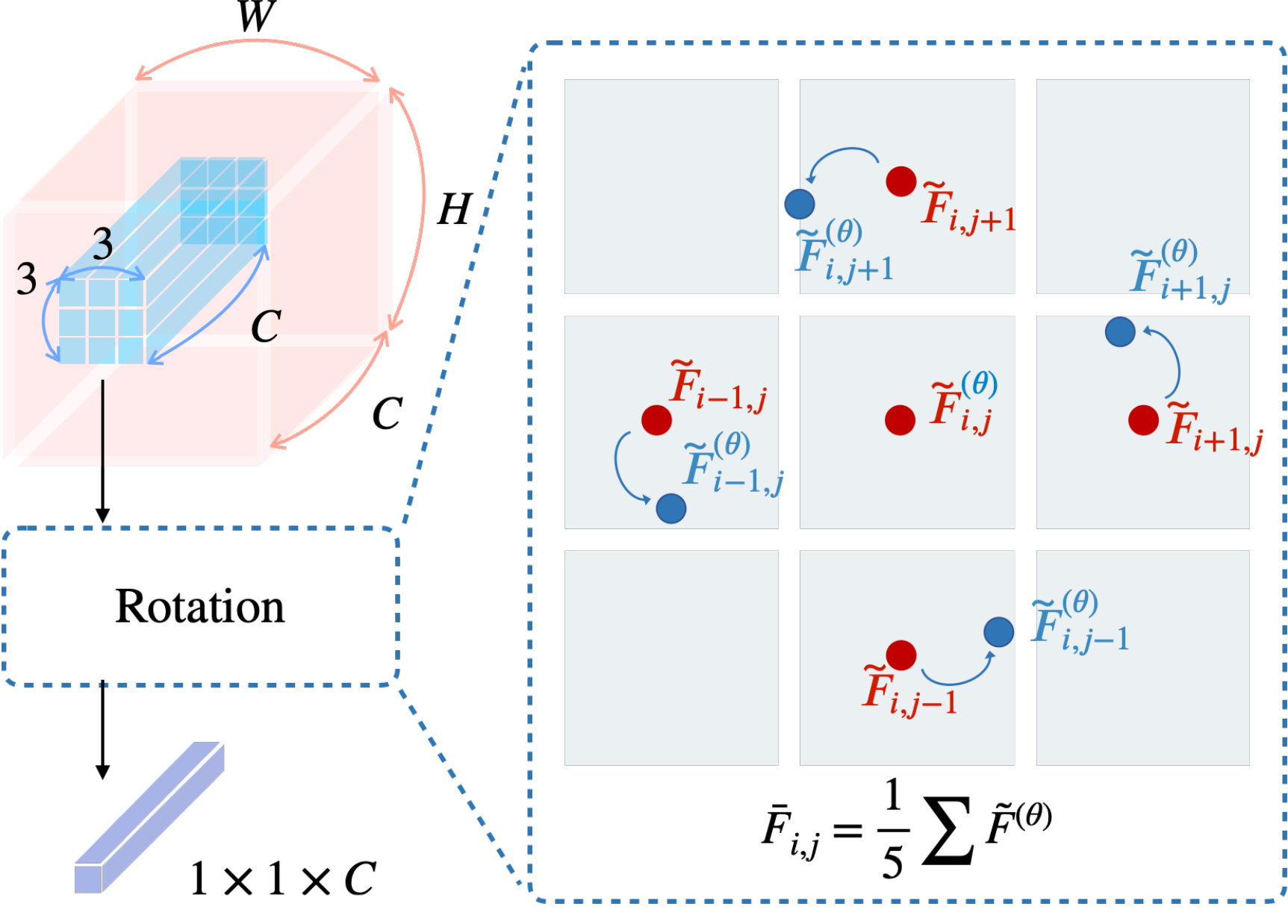}
  \caption{\textbf{Illustration of the rotation}. Each patch is rotated independently without changing the patch index. A rotated patch feature is obtained based on the features of its surrounding patches. Red: original patch features, Blue: rotated patch features.}
  \label{fig: rotation_method}
\end{figure}

Considering that we use 2D rotations to compensate for the difference of 3D camera poses, the rotation angles of different parts of the image should not be the same. And in order to prevent the rotation from changing the indices of the matching patches, we developed a local feature rotation algorithm to perform rotation on each patch feature locally. Since we do not know the appropriate rotation angle, we use gumbel softmax to select the best matched rotation angle. Assuming that adjacent patches should have similar features, that is, the feature vector of a patch $\tilde{F}_{i,j}$ can be approximated to the mean of its surrounding patches:
\begin{equation}
  \tilde{F}_{i,j}\approx \frac{1}{5}\left(\tilde{F}_{i,j}+\tilde{F}_{i-1,j}+\tilde{F}_{i+1,j}+\tilde{F}_{i,j-1}+\tilde{F}_{i,j+1}\right),
\end{equation}
where $\tilde{F}$ is the feature map of the image, and the index $0<=i<\frac{W}{8}, 0<=j<\frac{H}{8}$.

As shown in Fig.  \ref{fig: rotation_method}, given angle $\theta$, we define the rotated feature of patch $\tilde{F}_{i,j}$ by $\theta$ as
\begin{equation}
  \bar{F}_{i,j}^{(\theta)} = \frac{1}{5}\left(\tilde{F}_{i,j}^{(\theta)}+\tilde{F}_{i-1,j}^{(\theta)}+\tilde{F}_{i+1,j}^{(\theta)}+\tilde{F}_{i,j-1}^{(\theta)}+\tilde{F}_{i,j+1}^{(\theta)}\right),
\end{equation}
where
\begin{equation}
\begin{aligned}
  \tilde{F}_{i,j}^{(\theta)} = {} & \tilde{F}_{i,j}, \\
  \tilde{F}_{i+1,j}^{(\theta)} ={}& \tilde{F}(i+\cos{\theta},j+\sin{\theta}),\\
  \tilde{F}_{i,j+1}^{(\theta)} ={}& \tilde{F}(i+\cos{(\theta+\frac\pi2)},j+\sin{(\theta+\frac\pi2)}),\\
  \tilde{F}_{i-1,j}^{(\theta)} ={}& \tilde{F}(i+\cos{(\theta+\pi)},j+\sin{(\theta+\pi)}),\\
  \tilde{F}_{i,j-1}^{(\theta)} ={}& \tilde{F}(i+\cos{(\theta+\frac{3\pi}{2})},j+\sin{(\theta+\frac{3\pi}{2})})
\end{aligned}
\end{equation}
are the bilinear interpolations from its related features, which can be computed efficiently using \texttt{grid\_sample} function in PyTorch.

Since we do not know the ground truth rotation angle, gumbel softmax \cite{jang2016categorical} is used in subsequent coarse matching module. In implementation we use $\theta \in \{0^{\circ}, 30^{\circ}\}$.

\subsection{OE Module}
\label{sec:OE Module}
After obtaining the coarse-level patch matching, we use the OE Module to implement fine matching, as shown in Fig.  \ref{fig:ppl}. For points in visible patches, we calculate the offset from the center of the matched patch using the heatmap produced by a small LoFTR module \cite{sun2021loftr}. For points in occluded patches, the position of the matching point is inferred based on the visible contents around it, which are obtained from the feature extraction, and local 3D occupancy.

The OE Module is used to compute the exact matching points position. First, we design a 3D occupancy estimation module. For each image, we obtain two tensors $O^F$ and $O^V$ via the feature extraction. $O^F$ is the tensor of dimension $C \times \frac{1}{2}H \times \frac{1}{2}W \times 1$, where C is the number of channels. $O^V$ is the tensor of dimension  $1 \times \frac{1}{2}H \times \frac{1}{2}W \times D$, where $D$ is the resolution along the depth. We calculate a 4D tensor as the outer product of $O^F$ and $O^V$ to represent the estimated 3D occupancy,

\begin{equation}
O = \mathrm{softmax}(O^F \otimes O^V).
\end{equation}
Based on the ground truth depths and camera poses, we compute the ground truth 3D occupancy as supervision. We will explain the 3D occupancy loss in Sec. \ref{sec:loss}. 

We then combine the estimated 3D occupancy and fine features via an attention module. Finally, based on the local features and 3D occupancy of both images, we can infer the specific location of each matching point.

There are two reasons why coarse-to-fine structure is used and why the 3D occupancy estimation is used only in fine structure: 1. The coarse-to-fine structure has been shown effective for matching between visible points. 2. 3D occupancy estimation by monocular models is not accurate, while inferring obscured points is difficult. A coarse-to-fine structure is used, while the fine feature is combined with the 3D occupancy, so that the 3D information estimated from the 3D occupancy can be used to infer the location of the occluded points. Thanks to coarse matching, the error of the inference can be controlled within certain bounds.

For example, as shown in Fig. \ref{fig:rotation}, we try to match the green boxes in the left and right images. The features of the matching points should be consistent with 3D occupancy. We can infer from 3D occupancy that the arrow in the figure indicates a fine matching pair.

\subsection{Matching and Loss}
\label{sec:loss}
\textbf{Coarse-Matching Loss} 
Our network treats the following as valid matches: visible-visible patches, visible-occluded patches, and occluded-visible patches. 

Based on the ground truth camera poses and depths, we compute the reprojection of the left image patch. When the projected depth of a patch is larger by a large margin than the corresponding depth of the right image, we treat the patch as being obscured by the object in front of it in the right image. We define the ratio of occluded points to the total number of pixels of the image as the occlusion ratio. 

We obtain three sets of ground truth patch matching: visible-visible matching $\mathcal{M}_{c-gt}^{(v,v)}$, visible-occluded matching $\mathcal{M}_{c-gt}^{(v,o)}$, and occluded-visible matching $\mathcal{M}_{c-gt}^{(o,v)}$.

We use dual-softmax \cite{rocco2018neighbourhood} to compute the matching probability for matches, following LoFTR \cite{sun2021loftr}. Formally, the confidence matrix $\mathcal{P}_{AB}$ is obtained by
\begin{equation}
 \mathcal{P}_{AB} = \mathrm{softmax}\big(\mathcal{S} (i, {}\cdot{})\big)_j \cdot \mathrm{softmax} \big(\mathcal{S} ({}\cdot{},j)\big)_i,
\end{equation}
where $\mathcal{S}$ is the score matrix calculated by
\begin{equation}
 \mathcal{S}(i,j) = \frac{1}{\tau}\langle \bar{F}_{i}^{A}, \bar{F}_{j}^{B} \rangle.
\end{equation}

Considering local feature rotation, both the left and right images have been rotated by $0^{\circ}$ and $\theta$. Therefore, we obtain three confidence matrices: 
\begin{equation}
\mathcal{P}_{AB}^{(0^{\circ}, 0^{\circ})}, \mathcal{P}_{AB}^{(\theta, 0^{\circ})},\mathcal{P}_{AB}^{(0^{\circ}, \theta)}.
\end{equation}
Note that rotating both images by $\theta$ is equivalent to no rotation. The gumbel softmax is then applied to the three confidence matrices to select the best matched rotation angles:
\begin{multline}
 \hat{\mathcal{P}}_{AB} =\\ \mathrm{gumbel\_softmax}\big(
 \mathcal{P}_{AB}^{(0^{\circ}, 0^{\circ})}, \mathcal{P}_{AB}^{(\theta, 0^{\circ})},\mathcal{P}_{AB}^{(0^{\circ}, \theta)}\big).
\end{multline}

We minimize the sum of loss over the three cases:
\begin{equation} 
\begin{aligned}
 \mathcal{L}_c =& \gamma(\mathcal{M}_{c-gt}^{(v,v)}, \hat{\mathcal{P}}_{AB}) \\
 &+ \lambda_1\gamma(\mathcal{M}_{c-gt}^{(v,o)}, \hat{\mathcal{P}}_{AB}) + \lambda_1\gamma(\mathcal{M}_{c-gt}^{(o,v)}, \hat{\mathcal{P}}_{AB}),
\end{aligned} 
\end{equation}
where $\gamma$ is the negative log-likelihood loss used in LoFTR \cite{sun2021loftr}:
\begin{equation}
 \gamma(\mathcal{M}, \mathcal{P}) = -\frac{1}{\big|\mathcal{M}\big|}\sum_{(i,j)\in \mathcal{M}}{\log{\mathcal{P}(i,j)}}.
\end{equation}

\textbf{3D occupancy Loss}
To generate ground truth 3D occupancy from ground truth depth and camera pose, we use the following steps:
 \begin{itemize} 
 \item Create a ground truth point cloud by combining the depths and camera poses of both images.
\item Convert the point cloud into a voxel representation by dividing the 3D space into small cubic cells and assigning each cell a value based on its 3D occupancy.
\item Project the voxel representation onto the current image by using the camera pose as the world coordinate system origin and setting the depth resolution to 64 voxels.
\end{itemize}

The 3D occupancy loss is:
\begin{equation}
 \mathcal{L}_{o} = \|{O - O^{gt}}\|_1,
\end{equation}
where $O$ is the estimated 3D occupancy, $O^{gt}$ is the ground truth 3D occupancy.

\textbf{Fine Matching Loss} 
The target is to optimize the weighted loss function proposed by LoFTR \cite{sun2021loftr}:
\begin{equation}
 \mathcal{L}_{f} = \mathcal{L}_{f}^{(v,v)} + \lambda_2 \mathcal{L}_{f}^{(v,o)} + \lambda_2 \mathcal{L}_{f}^{(o,v)},
\end{equation}
which is calculated by reprojection error.

\textbf{Total Loss}
The final loss function consists of the coarse-level loss, the fine-level loss, and the 3D occupancy loss:
\begin{equation}
 \mathcal{L} = \mathcal{L}_{c} + \lambda_3 \mathcal{L}_{f} + \lambda_4 \mathcal{L}_{o}.
\end{equation}

\subsection{Implementation Details}
\label{sec: details}
We used AdamW as the optimizer with a learning rate of 6e-3. The images are resized to $640\times480$ before feeding them into the network. We set the $\lambda_1$, $\lambda_2$, $\lambda_3$, and $\lambda_4$ to 1.0, 1.0, 1.0, and 0.1, respectively, following the method proposed by Sec \ref{sec:method}. These parameters are chosen to balance the accuracy and efficiency of image matching.

\section{Experiments}
\label{sec:Experiments}
\subsection{Datasets}
We validate the effectiveness of our method through the experiments on real-world dataset ScanNet and simulated dataset TartanAir. The MegaDepth \cite{li2018megadepth} dataset is also commonly used for image matching, but it's not adopted here due to its relatively large depth map error. Because our method heavily relies on the ground truth depth and occupancy, we believe that our algorithm is not suitable for training on MegaDepth. TartanAir is a simulated dataset that offers accurate camera pose and depth information and includes outdoor scenes like MegaDepth. We use ScanNet and TartanAir dataset to demonstrate the generalization and adaptability of the algorithm in the case of inexact depth, as well as its effectiveness in both indoor and outdoor scenarios.

\textbf{ScanNet} We use ScanNet \cite{dai2017scannet} to demonstrate the effectiveness of our method for pose estimation. ScanNet is an RGB-D video dataset containing 2.5 million views from more than 1500 scans, annotated with 3D camera poses, surface reconstructions, and instance-level semantic segmentations.

Following the evaluation procedure of SuperGlue \cite{sarlin2020superglue}, we sample 230M image pairs for training with overlap scores between 0.4 and 0.8. The overlap score of an image pair is defined by reprojecting one image to the other using the ground truth depths and poses, and calculating the proportion of pixels that are not out of bounds after reprojection. We evaluate our method on the test set of 1500 image pairs. All images and depth maps are resized to $640\times480$. 

\textbf{TartanAir} TartanAir \cite{wang2020tartanair} is a challenging synthetic benchmark for evaluating SLAM algorithms. The dataset contains both indoor and outdoor scenes, covering a large variety of scenes and motion patterns. Data was collected in a photorealistic simulation environment in the presence of a variety of light conditions, weather, and moving objects. Unlike ScanNet, which uses a physical data collection platform to collect data, the depth and pose in the TartanAir dataset are completely exact.
We build the test set using the same test split as Droid-SLAM \cite{teed2021droid}, which contains 32 scenes. In each scenario, we randomly selected 50 image pairs satisfying an overlap score between 0.4 and 0.8 and an occlusion ratio larger than 0.3, resulting in a test set of 1600 image pairs.

\subsection{Pose Estimation}
\label{sec:scannet}
\begin{table}
\begin{center}
\begin{tabular}{l|ccc}
\hline
  \textbf{Method(ScanNet)} & \multicolumn{3}{c}{Pose estimation AUC\%}\\
  \cline{2-4}
   & $@5^{\circ}$ & $@10^{\circ}$ & $@20^{\circ}$\\
\hline\hline
  SuperPoint &7.6 & 17.1 & 28.8 \\
  SuperPoint + SuperGlue & 16.2 & 33.8 & 51.8 \\
  LoFTR & 22.1 & 40.8 & 57.6 \\
  QuadTree LoFTR & 24.9 & 44.7 & 61.8\\
  ASpanFormer & 25.6 & 46.0 & 63.3\\
  \textbf{Occ$^2$Net-s} & \textbf{25.0} & \textbf{46.1} & \textbf{64.6}\\  
  \textbf{Occ$^2$Net-l} & \textbf{28.0} & \textbf{48.7} & \textbf{66.9}\\
\hline\hline
  \textbf{Method(TartanAir)} & \multicolumn{3}{c}{Pose estimation AUC\%}\\
\cline{2-4}
   & $@5^{\circ}$ & $@10^{\circ}$ & $@20^{\circ}$\\
\hline\hline
  SuperPoint & 13.1 & 27.9 & 30.0\\
  SuperPoint + SuperGlue & 17.3 & 24.8 & 32.5\\
  LoFTR & 17.6 & 26.2 & 34.1\\
  QuadTree LoFTR & 18.3 & 28.6 & 34.7\\
  \textbf{Occ$^2$Net-s} & \textbf{20.1} & \textbf{32.2} & \textbf{43.4}\\
  \textbf{Occ$^2$Net-l} & \textbf{28.0} & \textbf{39.4} & \textbf{47.3}\\

\hline
\end{tabular}
\end{center}
\caption{Evaluation. The percentage AUC of pose error is reported. Our approach outperforms state-of-the-art methods.}
\label{tab: results}
\end{table}

\begin{figure*}[t]
 \centering
  \includegraphics[width=1.0\linewidth]{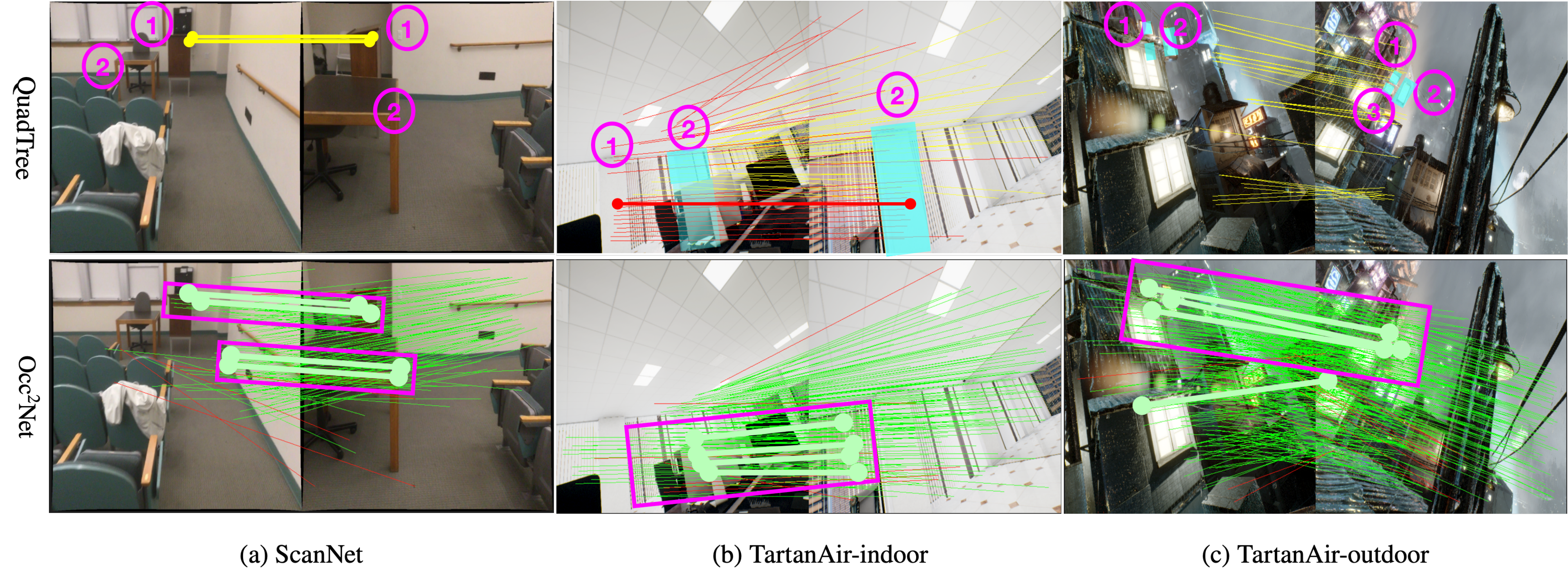}
  \caption{\textbf{Matching examples of ScanNet, TartanAir-indoor, and TartanAir-outdoor}. We compare the QuadTree LoFTR \cite{tang2022quadtree} with our results. The green and yellow lines are proper matches by QuadTree LoFTR and Occ$^2$Net, respectively, while the red lines are false matches (error larger than 10 pixels). We highlight some matched pairs with the purple boxes. The blue mask with the same label indicates the correct matching region. The left and right images in (a) include different sides of lectern \ding{172} and Chair \ding{173}, respectively. The left image in (b) contains walls \ding{172} and \ding{173}, with Wall \ding{173} partially obscured by the display. The right image in (b) contains only wall \ding{173}. The left image in (c) contains only lights \ding{172} and \ding{173}, while the right image contains lights \ding{172}, \ding{173} and \ding{174}.}
  \label{fig:result}
\end{figure*}

\textbf{Evaluation protocol} Following SuperGlue \cite{sarlin2020superglue}, we report the percentage AUC of the pose error at thresholds ($5^{\circ}, 10^{\circ}, 20^{\circ}$), where the pose error is defined as the maximum of angular error in the rotation and displacement error in the translation. To recover the camera poses, we solve the essential matrix from predicted matches using RANSAC \cite{derpanis2010overview}. Note that the occluded points also lie on the epipolar plane of the camera rays, which results in exactly the same equations as used for traditional visible-visible matching.

\textbf{Results of pose estimation}
As shown in Tab. \ref{tab: results}, Our method achieves the best performance in pose estimation accuracy compared to all competitors on both ScanNet and TartanAir. This shows that our method is effective for indoor and outdoor, real and simulated scenarios.  Occ$^2$Net-s, which is a truncated version of  Occ$^2$Net-l, does not distinct between visible and occluded points in the attention component, and use only one confidence matrix instead of three. The Occ$^2$Net-s has fewer parameters than QuadTree LoFTR \cite{tang2022quadtree}. 

The shared weight has a certain impact on the final performance. As we explained before, the exact location of the matching point with occlusion is difficult to estimate, so the increase of $@5^{\circ}$ is not obvious. However, the model without shared weight is equivalent to adding the occluded matching points with visible matching points. The two kinds of matching will not affect each other, and the accuracy improvement is obvious.

The TartanAir test set contains more image pairs whose occlusion ratio is larger than $40\%$, so the matching effect improves even more.

\subsection{The effect of occlusion on pose estimation}

\begin{figure}[t]
 \centering
  \includegraphics[width=1.0\linewidth]{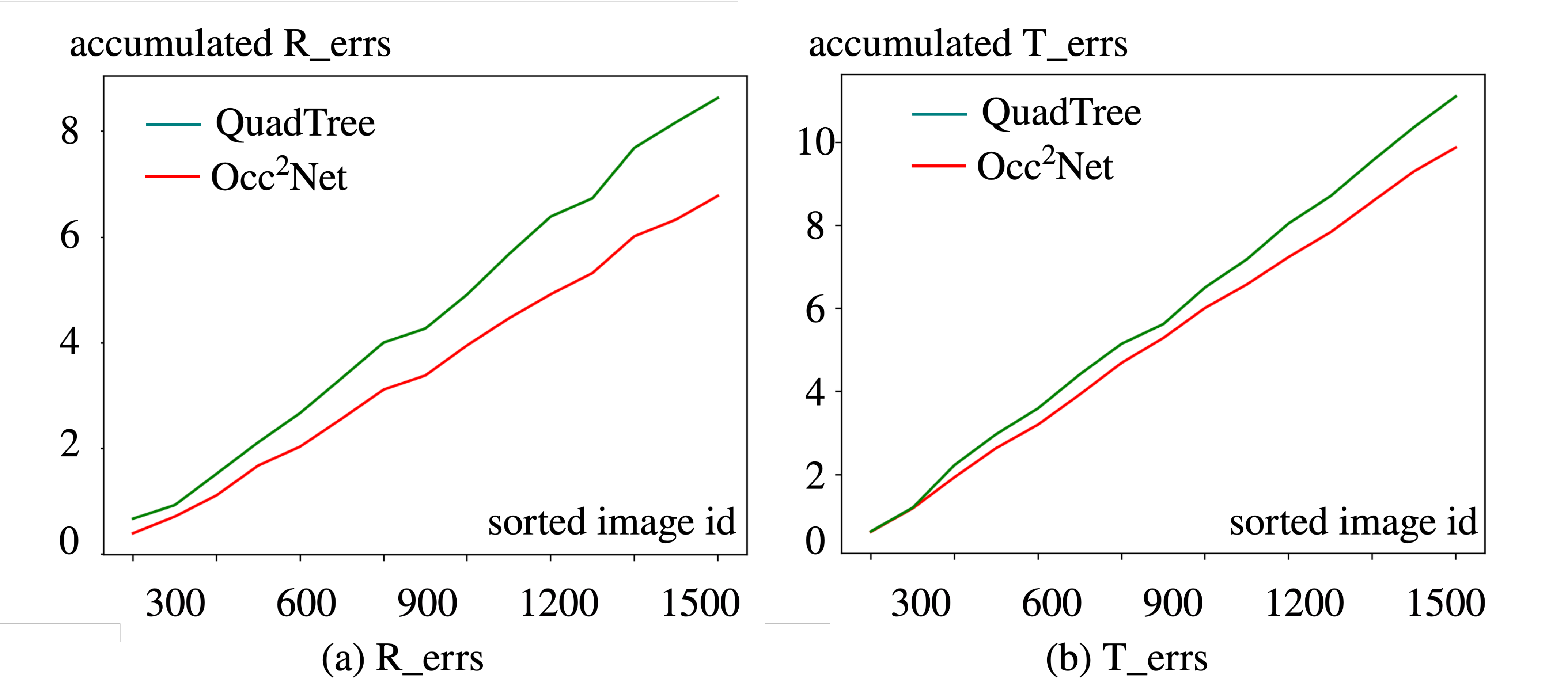}
  \caption{\textbf{Curves of the cumulative average pose error and the cumulative amount of images} (as the occlusion ratio gradually increases). }
  \label{fig:occlusion}
\end{figure}

\textbf{Quantitatively} To quantitatively analyze the effect of occlusion on pose estimation, we compute the occlusion ratio for all image pairs in the ScanNet test set. The occlusion ratio is defined as the ratio of pixels that are visible in one image but occluded in another. We sort the test set according to the occlusion ratio, and plot the function of the cumulative average pose error and the cumulative amount of images, showing the relation of the mean estimation error and the occlusion ratio. As shown in Fig. \ref{fig:occlusion}, our algorithm has constantly low pose error on image pairs with low or high occlusion ratio, while other algorithms fails more often on challenging cases.

\textbf{Qualitatively} Fig. \ref{fig:result} shows some examples of test datasets compared with QuadTree LoFTR \cite{tang2022quadtree}. The green and yellow lines are both properly matched, while the red lines are false matches. We highlight some matched pairs with the purple boxes. The first pair of images is difficult to match due to the excessive disparity. It is difficult for other algorithms to obtain the correct matching points, and even insufficient matches for pose estimation. In addition to adding matching points for low texture regions such as walls, our algorithm also successfully matches lectures \ding{172}, and chair \ding{173} with occluded ground, as shown by the purple boxes in Fig. \ref{fig:result}.

The second pair of images in Fig. \ref{fig:result} shows our algorithm has a better understanding of the spatial structure, thus more correct matching points are obtained. Occ$^2$Net can correctly match points that are occluded by the monitor, as shown in the purple box. The two walls labeled \ding{173} are correct matches, while the QuadTree LoFTR has many false matches between walls \ding{172} and \ding{173}. Although our algorithm has a small number of points on wall \ding{172} that match the curtain, these points are within the acceptable margin of error.

The third pair of images in Fig. \ref{fig:result} shows an outdoor scene with multiple occlusions among buildings. Our algorithm can obtain more matching point pairs, especially the matching relationship represented by the red mask in the purple box. Our method matches not only the two visible lights, but also the occluded light \ding{174}.

\subsection{Ablation Study}
To adequately understand the different modules in our method, we evaluate several variants in ScanNet with results shown in Tab. \ref{tab:ablation}. We use QuadTree LoFTR \cite{tang2022quadtree} as baseline.

\begin{table}
\begin{center}
\begin{tabular}{l|ccc}
\hline
  \textbf{Experiments(Occlusion)}  & \multicolumn{3}{c}{Pose estimation AUC\%} \\
  \cline{2-4}
   & $@5^{\circ}$ & $@10^{\circ}$ & $@20^{\circ}$ \\
\hline\hline
  1) Baseline & 24.9 & 44.7 & 61.8 \\
  2) Baseline $+$ Occlusion loss & 24.3 & 44.6 & 62.1\\
  3)  Occ$^2$Net $-$ Occlusion loss & 24.4 & 45.4 & 63.3 \\
  4)  Occ$^2$Net (visible-visible only) & 24.3 & 45.2 & 63.2 \\
\hline\hline  
  \textbf{Experiments(Module)}  & \multicolumn{3}{c}{Pose estimation AUC\%} \\
  \cline{2-4}
   & $@5^{\circ}$ & $@10^{\circ}$ & $@20^{\circ}$ \\
\hline\hline
  5) Baseline $+$ Feature extraction & 22.6 & 41.7 & 58.9 \\
  6) Exp. 4) $+$ Occlusion loss & 24.8 & 45.1 & 63.7 \\
  7) Baseline $+$ OA$\_$Module & 25.1	& 46.0 & 64.0 \\
  8)  Occ$^2$Net $-$ Occupancy & 25.7 & 46.4 & 64.2 \\
\hline\hline  
  9) \textbf{Occ$^2$Net} & \textbf{28.0} & \textbf{48.7} & \textbf{66.9}\\
  
\hline
\end{tabular}
\end{center}
\caption{Experiments of different variants: 1) Baseline: QuadTree LoFTR \cite{tang2022quadtree}.
2) Baseline+Occlusion loss: only adding visible-occluded matches for supervision;
3)  Occ$^2$Net-Occlusion loss: Using  Occ$^2$Net without occlusion supervised;
4)  Occ$^2$Net(visible-visible only): Matching between only visible points;
5) Baseline+feature extraction: using our feature extraction backbone, instead of LoFTR feature extraction backbone;
6) Exp. 4)+Occlusion loss: using our feature extraction backbone and adding visible-occluded matches for supervision;
7) Baseline+OA$\_$Module: adding Rotation Alignment method;
8)  Occ$^2$Net-Occupancy:  Occ$^2$Net without 3D occupancy estimation;
9)  Occ$^2$Net: our method.}
\label{tab:ablation}
\end{table}

As shown in Tab. \ref{tab:ablation} 1) and 2), only adding supervision for visible-occluded points isn't useful. In order to realize the matching of visible points and occluded points, it is necessary to design a network structure that can be aware of occlusion and infer the occluded position.

Comparing 4) and 9), pose estimation using only visible-visible points significantly gets worse result compared to using both occluded and visible points.

Comparing 1) and 5), our feature extraction provides tensors for 3D occupancy estimation and has fewer parameters than QuadTree LoFTR's feature extraction, which leads to poor matching effect of visible points. 
Comparing 5) and 6), our feature extraction is valid for occluded point matching.

Comparing 9) and 8), the 3D occupancy estimation improves the occluded point matching with occlusion supervision. Furthermore, by matching occluded and visible points simultaneously, it also enhances the visible-visible matching and thus refines the overall matching performance.

Note that all experiments in Tab. \ref{tab:ablation} cannot be directly compared with Occ$^2$Net-s  because they contain different number of parameters.

\section{Conclusion}
We introduce a way of matching images that takes into account the points that are occluded by other objects. We design a network structure, Occ$^2$Net, that can be aware of the existence of occluded points to some degree.  Occ$^2$Net uses feature extraction to get multi-scale global and positional features that help guess the occlusion information. The OA Module uses attention and Rotation Alignment, which are useful for getting more correct matching pairs in the later coarse-to-fine process. The OE Module uses 3D occupancy estimation to combine fine features for fine matching. Experiments show that our method achieves this goal and greatly improves the accuracy of pose estimation. We hope that our work paves the way for future researchers to explore better image matching for occlusion scenarios.

\textbf{Limitations} Although our network can identify the occluded points and match them with the visible points, it is hard to find out the exact location of these hidden points. Even though we first suggested the idea of a ray made up of visible and hidden points for image matching, we degenerate the ray to only two points because of data limitations. In the future, more complex scenarios involving multiple layers of occlusion should be considered.

{\small
\bibliographystyle{ieee_fullname}
\bibliography{egbib}

\begin{thebibliography}{10}\itemsep=-1pt

\bibitem{ao2022image}
Jiayang Ao, Krista~A Ehinger, and Qiuhong Ke.
\newblock Image amodal completion: A survey.
\newblock {\em arXiv preprint arXiv:2207.02062}, 2022.

\bibitem{back2022unseen}
Seunghyeok Back, Joosoon Lee, Taewon Kim, Sangjun Noh, Raeyoung Kang, Seongho
  Bak, and Kyoobin Lee.
\newblock Unseen object amodal instance segmentation via hierarchical occlusion
  modeling.
\newblock In {\em 2022 International Conference on Robotics and Automation
  (ICRA)}, pages 5085--5092. IEEE, 2022.

\bibitem{bay2008speeded}
Herbert Bay, Andreas Ess, Tinne Tuytelaars, and Luc Van~Gool.
\newblock Speeded-up robust features (surf).
\newblock {\em Computer vision and image understanding}, 110(3):346--359, 2008.

\bibitem{campos2021orb}
Carlos Campos, Richard Elvira, Juan J~G{\'o}mez Rodr{\'\i}guez, Jos{\'e}~MM
  Montiel, and Juan~D Tard{\'o}s.
\newblock Orb-slam3: An accurate open-source library for visual,
  visual--inertial, and multimap slam.
\newblock {\em IEEE Transactions on Robotics}, 37(6):1874--1890, 2021.

\bibitem{cao2022monoscene}
Anh-Quan Cao and Raoul de Charette.
\newblock Monoscene: Monocular 3d semantic scene completion.
\newblock In {\em Proceedings of the IEEE/CVF Conference on Computer Vision and
  Pattern Recognition}, pages 3991--4001, 2022.

\bibitem{chen2022aspanformer}
Hongkai Chen, Zixin Luo, Lei Zhou, Yurun Tian, Mingmin Zhen, Tian Fang, David
  Mckinnon, Yanghai Tsin, and Long Quan.
\newblock Aspanformer: Detector-free image matching with adaptive span
  transformer.
\newblock In {\em European Conference on Computer Vision}, pages 20--36.
  Springer, 2022.

\bibitem{dai2017scannet}
Angela Dai, Angel~X Chang, Manolis Savva, Maciej Halber, Thomas Funkhouser, and
  Matthias Nie{\ss}ner.
\newblock Scannet: Richly-annotated 3d reconstructions of indoor scenes.
\newblock In {\em Proceedings of the IEEE conference on computer vision and
  pattern recognition}, pages 5828--5839, 2017.

\bibitem{danielsson1980euclidean}
Per-Erik Danielsson.
\newblock Euclidean distance mapping.
\newblock {\em Computer Graphics and image processing}, 14(3):227--248, 1980.

\bibitem{derpanis2010overview}
Konstantinos~G Derpanis.
\newblock Overview of the ransac algorithm.
\newblock {\em Image Rochester NY}, 4(1):2--3, 2010.

\bibitem{detone2017toward}
Daniel DeTone, Tomasz Malisiewicz, and Andrew Rabinovich.
\newblock Toward geometric deep slam.
\newblock {\em arXiv preprint arXiv:1707.07410}, 2017.

\bibitem{detone2018superpoint}
Daniel DeTone, Tomasz Malisiewicz, and Andrew Rabinovich.
\newblock Superpoint: Self-supervised interest point detection and description.
\newblock In {\em Proceedings of the IEEE conference on computer vision and
  pattern recognition workshops}, pages 224--236, 2018.

\bibitem{dong2022incremental}
Qiaole Dong, Chenjie Cao, and Yanwei Fu.
\newblock Incremental transformer structure enhanced image inpainting with
  masking positional encoding.
\newblock In {\em Proceedings of the IEEE/CVF Conference on Computer Vision and
  Pattern Recognition}, pages 11358--11368, 2022.

\bibitem{dusmanu2019d2}
Mihai Dusmanu, Ignacio Rocco, Tomas Pajdla, Marc Pollefeys, Josef Sivic,
  Akihiko Torii, and Torsten Sattler.
\newblock D2-net: A trainable cnn for joint description and detection of local
  features.
\newblock In {\em Proceedings of the ieee/cvf conference on computer vision and
  pattern recognition}, pages 8092--8101, 2019.

\bibitem{germain2021visual}
Hugo Germain, Vincent Lepetit, and Guillaume Bourmaud.
\newblock Visual correspondence hallucination.
\newblock {\em arXiv preprint arXiv:2106.09711}, 2021.

\bibitem{jang2016categorical}
Eric Jang, Shixiang Gu, and Ben Poole.
\newblock Categorical reparameterization with gumbel-softmax.
\newblock {\em arXiv preprint arXiv:1611.01144}, 2016.

\bibitem{jiang2021cotr}
Wei Jiang, Eduard Trulls, Jan Hosang, Andrea Tagliasacchi, and Kwang~Moo Yi.
\newblock Cotr: Correspondence transformer for matching across images.
\newblock In {\em Proceedings of the IEEE/CVF International Conference on
  Computer Vision}, pages 6207--6217, 2021.

\bibitem{lee2022instance}
Hyunmin Lee and Jaesik Park.
\newblock Instance-wise occlusion and depth orders in natural scenes.
\newblock In {\em Proceedings of the IEEE/CVF Conference on Computer Vision and
  Pattern Recognition}, pages 21210--21221, 2022.

\bibitem{li2021deep}
Guangqiang Li, Lei Yu, and Shumin Fei.
\newblock A deep-learning real-time visual slam system based on multi-task
  feature extraction network and self-supervised feature points.
\newblock {\em Measurement}, 168:108403, 2021.

\bibitem{li2022mat}
Wenbo Li, Zhe Lin, Kun Zhou, Lu Qi, Yi Wang, and Jiaya Jia.
\newblock Mat: Mask-aware transformer for large hole image inpainting.
\newblock In {\em Proceedings of the IEEE/CVF conference on computer vision and
  pattern recognition}, pages 10758--10768, 2022.

\bibitem{li2020dual}
Xinghui Li, Kai Han, Shuda Li, and Victor Prisacariu.
\newblock Dual-resolution correspondence networks.
\newblock {\em Advances in Neural Information Processing Systems},
  33:17346--17357, 2020.

\bibitem{li2018megadepth}
Zhengqi Li and Noah Snavely.
\newblock Megadepth: Learning single-view depth prediction from internet
  photos.
\newblock In {\em Proceedings of the IEEE conference on computer vision and
  pattern recognition}, pages 2041--2050, 2018.

\bibitem{lowe2004distinctive}
David~G Lowe.
\newblock Distinctive image features from scale-invariant keypoints.
\newblock {\em International journal of computer vision}, 60(2):91--110, 2004.

\bibitem{luo2019contextdesc}
Zixin Luo, Tianwei Shen, Lei Zhou, Jiahui Zhang, Yao Yao, Shiwei Li, Tian Fang,
  and Long Quan.
\newblock Contextdesc: Local descriptor augmentation with cross-modality
  context.
\newblock In {\em Proceedings of the IEEE/CVF conference on computer vision and
  pattern recognition}, pages 2527--2536, 2019.

\bibitem{ma2018shufflenet}
Ningning Ma, Xiangyu Zhang, Hai-Tao Zheng, and Jian Sun.
\newblock Shufflenet v2: Practical guidelines for efficient cnn architecture
  design.
\newblock In {\em Proceedings of the European conference on computer vision
  (ECCV)}, pages 116--131, 2018.

\bibitem{miao2023occdepth}
Ruihang Miao, Weizhou Liu, Mingrui Chen, Zheng Gong, Weixin Xu, Chen Hu, and
  Shuchang Zhou.
\newblock Occdepth: A depth-aware method for 3d semantic scene completion.
\newblock {\em arXiv preprint arXiv:2302.13540}, 2023.

\bibitem{mildenhall2021nerf}
Ben Mildenhall, Pratul~P Srinivasan, Matthew Tancik, Jonathan~T Barron, Ravi
  Ramamoorthi, and Ren Ng.
\newblock Nerf: Representing scenes as neural radiance fields for view
  synthesis.
\newblock {\em Communications of the ACM}, 65(1):99--106, 2021.

\bibitem{mohan2022amodal}
Rohit Mohan and Abhinav Valada.
\newblock Amodal panoptic segmentation.
\newblock In {\em Proceedings of the IEEE/CVF Conference on Computer Vision and
  Pattern Recognition}, pages 21023--21032, 2022.

\bibitem{mur2015orb}
Raul Mur-Artal, Jose Maria~Martinez Montiel, and Juan~D Tardos.
\newblock Orb-slam: a versatile and accurate monocular slam system.
\newblock {\em IEEE transactions on robotics}, 31(5):1147--1163, 2015.

\bibitem{mur2017orb}
Raul Mur-Artal and Juan~D Tard{\'o}s.
\newblock Orb-slam2: An open-source slam system for monocular, stereo, and
  rgb-d cameras.
\newblock {\em IEEE transactions on robotics}, 33(5):1255--1262, 2017.

\bibitem{norouzi2012hamming}
Mohammad Norouzi, David~J Fleet, and Russ~R Salakhutdinov.
\newblock Hamming distance metric learning.
\newblock {\em Advances in neural information processing systems}, 25, 2012.

\bibitem{qin2020sp}
Zixuan Qin, Mengxiao Yin, Guiqing Li, and Feng Yang.
\newblock Sp-flow: Self-supervised optical flow correspondence point prediction
  for real-time slam.
\newblock {\em Computer Aided Geometric Design}, 82:101928, 2020.

\bibitem{quan2022image}
Weize Quan, Ruisong Zhang, Yong Zhang, Zhifeng Li, Jue Wang, and Dong-Ming Yan.
\newblock Image inpainting with local and global refinement.
\newblock {\em IEEE Transactions on Image Processing}, 31:2405--2420, 2022.

\bibitem{rocco2020efficient}
Ignacio Rocco, Relja Arandjelovi{\'c}, and Josef Sivic.
\newblock Efficient neighbourhood consensus networks via submanifold sparse
  convolutions.
\newblock In {\em European conference on computer vision}, pages 605--621.
  Springer, 2020.

\bibitem{rocco2018neighbourhood}
Ignacio Rocco, Mircea Cimpoi, Relja Arandjelovi{\'c}, Akihiko Torii, Tomas
  Pajdla, and Josef Sivic.
\newblock Neighbourhood consensus networks.
\newblock {\em Advances in neural information processing systems}, 31, 2018.

\bibitem{rocco2020ncnet}
Ignacio Rocco, Mircea Cimpoi, Relja Arandjelovic, Akihiko Torii, Tomas Pajdla,
  and Josef Sivic.
\newblock Ncnet: Neighbourhood consensus networks for estimating image
  correspondences.
\newblock {\em IEEE Transactions on Pattern Analysis and Machine Intelligence},
  2020.

\bibitem{rublee2011orb}
Ethan Rublee, Vincent Rabaud, Kurt Konolige, and Gary Bradski.
\newblock Orb: An efficient alternative to sift or surf.
\newblock In {\em 2011 International conference on computer vision}, pages
  2564--2571. Ieee, 2011.

\bibitem{sarlin2020superglue}
Paul-Edouard Sarlin, Daniel DeTone, Tomasz Malisiewicz, and Andrew Rabinovich.
\newblock Superglue: Learning feature matching with graph neural networks.
\newblock In {\em Proceedings of the IEEE/CVF conference on computer vision and
  pattern recognition}, pages 4938--4947, 2020.

\bibitem{senoussaoui2013study}
Mohammed Senoussaoui, Patrick Kenny, Themos Stafylakis, and Pierre Dumouchel.
\newblock A study of the cosine distance-based mean shift for telephone speech
  diarization.
\newblock {\em IEEE/ACM Transactions on Audio, Speech, and Language
  Processing}, 22(1):217--227, 2013.

\bibitem{shi2020point}
Weijing Shi and Raj Rajkumar.
\newblock Point-gnn: Graph neural network for 3d object detection in a point
  cloud.
\newblock In {\em Proceedings of the IEEE/CVF conference on computer vision and
  pattern recognition}, pages 1711--1719, 2020.

\bibitem{sun2021loftr}
Jiaming Sun, Zehong Shen, Yuang Wang, Hujun Bao, and Xiaowei Zhou.
\newblock Loftr: Detector-free local feature matching with transformers.
\newblock In {\em Proceedings of the IEEE/CVF conference on computer vision and
  pattern recognition}, pages 8922--8931, 2021.

\bibitem{tang2022quadtree}
Shitao Tang, Jiahui Zhang, Siyu Zhu, and Ping Tan.
\newblock Quadtree attention for vision transformers.
\newblock {\em arXiv preprint arXiv:2201.02767}, 2022.

\bibitem{teed2021droid}
Zachary Teed and Jia Deng.
\newblock Droid-slam: Deep visual slam for monocular, stereo, and rgb-d
  cameras.
\newblock {\em Advances in Neural Information Processing Systems},
  34:16558--16569, 2021.

\bibitem{tyszkiewicz2020disk}
Micha{\l} Tyszkiewicz, Pascal Fua, and Eduard Trulls.
\newblock Disk: Learning local features with policy gradient.
\newblock {\em Advances in Neural Information Processing Systems},
  33:14254--14265, 2020.

\bibitem{vaswani2017attention}
Ashish Vaswani, Noam Shazeer, Niki Parmar, Jakob Uszkoreit, Llion Jones,
  Aidan~N Gomez, {\L}ukasz Kaiser, and Illia Polosukhin.
\newblock Attention is all you need.
\newblock {\em Advances in neural information processing systems}, 30, 2017.

\bibitem{wang2020tartanair}
Wenshan Wang, Delong Zhu, Xiangwei Wang, Yaoyu Hu, Yuheng Qiu, Chen Wang, Yafei
  Hu, Ashish Kapoor, and Sebastian Scherer.
\newblock Tartanair: A dataset to push the limits of visual slam.
\newblock In {\em 2020 IEEE/RSJ International Conference on Intelligent Robots
  and Systems (IROS)}, pages 4909--4916. IEEE, 2020.

\bibitem{wu2021rethinking}
Kan Wu, Houwen Peng, Minghao Chen, Jianlong Fu, and Hongyang Chao.
\newblock Rethinking and improving relative position encoding for vision
  transformer.
\newblock In {\em Proceedings of the IEEE/CVF International Conference on
  Computer Vision}, pages 10033--10041, 2021.

\bibitem{xu2020cnn}
Zhilin Xu, Jincheng Yu, Chao Yu, Hao Shen, Yu Wang, and Huazhong Yang.
\newblock Cnn-based feature-point extraction for real-time visual slam on
  embedded fpga.
\newblock In {\em 2020 IEEE 28th Annual International Symposium on
  Field-Programmable Custom Computing Machines (FCCM)}, pages 33--37. IEEE,
  2020.

\bibitem{yi2016lift}
Kwang~Moo Yi, Eduard Trulls, Vincent Lepetit, and Pascal Fua.
\newblock Lift: Learned invariant feature transform.
\newblock In {\em European conference on computer vision}, pages 467--483.
  Springer, 2016.

\bibitem{yi2018learning}
Kwang~Moo Yi, Eduard Trulls, Yuki Ono, Vincent Lepetit, Mathieu Salzmann, and
  Pascal Fua.
\newblock Learning to find good correspondences.
\newblock In {\em Proceedings of the IEEE conference on computer vision and
  pattern recognition}, pages 2666--2674, 2018.

\bibitem{zeng2022aggregated}
Yanhong Zeng, Jianlong Fu, Hongyang Chao, and Baining Guo.
\newblock Aggregated contextual transformations for high-resolution image
  inpainting.
\newblock {\em IEEE Transactions on Visualization and Computer Graphics}, 2022.

\bibitem{zhang2019learning}
Jiahui Zhang, Dawei Sun, Zixin Luo, Anbang Yao, Lei Zhou, Tianwei Shen, Yurong
  Chen, Long Quan, and Hongen Liao.
\newblock Learning two-view correspondences and geometry using order-aware
  network.
\newblock In {\em Proceedings of the IEEE/CVF international conference on
  computer vision}, pages 5845--5854, 2019.

\bibitem{zhang2022gan}
Xian Zhang, Xin Wang, Canghong Shi, Zhe Yan, Xiaojie Li, Bin Kong, Siwei Lyu,
  Bin Zhu, Jiancheng Lv, Youbing Yin, et~al.
\newblock De-gan: Domain embedded gan for high quality face image inpainting.
\newblock {\em Pattern Recognition}, 124:108415, 2022.

\bibitem{zhong2018detect}
Fangwei Zhong, Sheng Wang, Ziqi Zhang, and Yizhou Wang.
\newblock Detect-slam: Making object detection and slam mutually beneficial.
\newblock In {\em 2018 IEEE Winter Conference on Applications of Computer
  Vision (WACV)}, pages 1001--1010. IEEE, 2018.

\end{thebibliography}
}

\end{document}